\documentclass[10pt,twocolumn,letterpaper]{article}

\usepackage{iccv}              
\usepackage{amsmath}
\usepackage{pifont}

\usepackage{amsmath}
\DeclareMathOperator*{\argmax}{arg\,max}
\usepackage{multirow}
\usepackage{makecell}

\definecolor{iccvblue}{rgb}{0.21,0.49,0.74}
\usepackage[pagebackref,breaklinks,colorlinks,allcolors=iccvblue]{hyperref}

\def\paperID{1646} 
\def\confName{ICCV}
\def\confYear{2025}

\title{SC-Captioner: Improving Image Captioning with Self-Correction by Reinforcement Learning}

\author{
Lin Zhang$^{1}$\thanks{Work was done when interned at StepFun. $^\dagger$Corresponding author.\quad\text{\ding{168}}Project leader.
} , \ 
Xianfang Zeng$^{2 \text{\ding{168}}}$, \ 
Kangcong Li$^1$, \ 
Gang Yu$^2$, \ 
Tao Chen$^{1,3}$$^\dagger$
 \\
$^1$ College of Future Information Technology, Fudan University \\
$^2$ StepFun \\
$^3$ Shanghai Innovation Institute \\
{\tt\small 22110720068@m.fudan.edu.cn, eetchen@fudan.edu.cn}
}

\begin{document}
\maketitle
\begin{abstract}

We propose SC-Captioner, a reinforcement learning framework that enables the self-correcting capability of image caption models. 
Our crucial technique lies in the design of the reward function to incentivize accurate caption corrections.
Specifically, the predicted and reference captions are decomposed into object, attribute, and relation sets using scene-graph parsing algorithms. 
%
We calculate the set difference between sets of initial and self-corrected captions to identify added and removed elements.
These elements are matched against the reference sets to calculate correctness bonuses for accurate refinements and mistake punishments for wrong additions and removals, thereby forming the final reward.
%
For image caption quality assessment, we propose a set of metrics refined from CAPTURE that alleviate its incomplete precision evaluation and inefficient relation matching problems.
Furthermore, we collect a fine-grained annotated image caption dataset, RefinedCaps, consisting of 6.5K diverse images from COCO dataset.
%
Experiments show that applying SC-Captioner on large visual-language models can generate better image captions across various scenarios, significantly outperforming the direct preference optimization training strategy. Our code is available at: \url{https://github.com/zl2048/SC-Captioner}

\end{abstract}    
\section{Introduction}
\label{sec:intro}

Image captioning is a fundamental task in computer vision and has been studied for years~\cite{Vinyals_2015_CVPR, liu2017improved, cornia2020meshed, liu2017attention, huang2019attention, wang2022git}. This technique is widely used in auto generation of film captions and data auto-labeling in recent text-to-image generation training. In recent years, large language models (LLMs) and large vision-language models (LVLMs) have emerged and injected new vitality into this field. LVLMs~\cite{li2023blip2,liu2024llava1.5,wang2024qwen2} can produce more precise and detailed image captions since they have larger parameter numbers and more extensive training data compared with traditional architectures. 

\begin{figure}[t]
  \centering
  \includegraphics[width=0.88\linewidth]{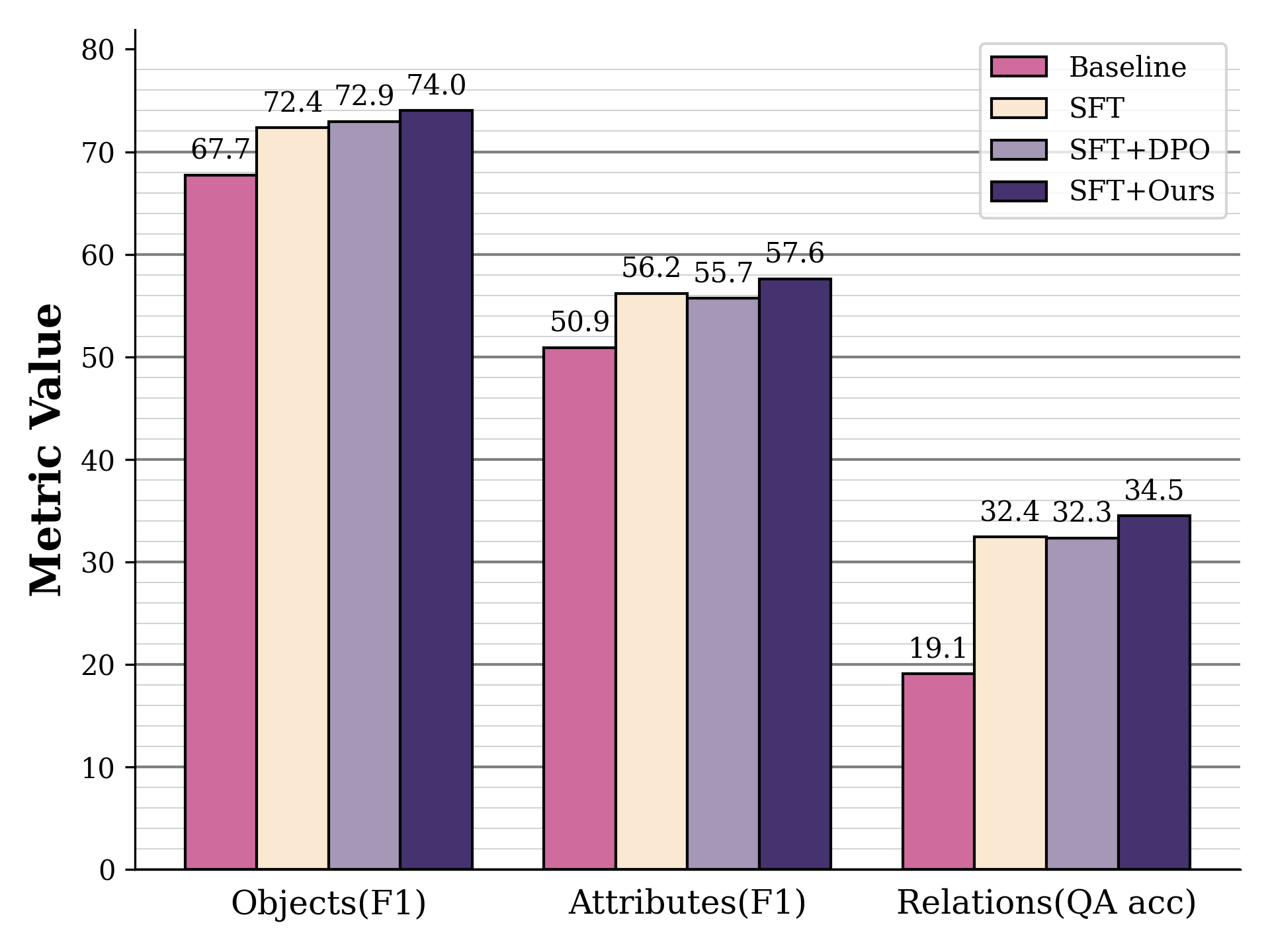}

   \caption{Comparisons of Qwen2-VL-7B baseline, Supervised Fine-Tuning (SFT), Direct Preference Optimization (DPO), and our method on DOCCI500. All models are trained on RefinedCaps.
   Our proposed method outperforms DPO and SFT baseline.
   }
   \label{fig:1}
\end{figure}

However, image captioning with LVLMs still faces two challanges which can be summarized as \textbf{precision} and \textbf{recall}. Precision refers to the proportion of the content in the generated caption that actually exists in the image. LVLMs sometimes give statements about contents not exists in the images. The phenomenon is also called ``hallucination". Recall measures the extent to which the generated caption captures all the relevant details and objects present in the image. While longer captions can achieve higher recall, they are more likely to contain hallucinations. A balance between precision and recall is necessary to avoid both missing important details and introducing incorrect information.

When generating image captions, providing humans with initial caption for post-editing is often more efficient than creating a caption from scratch. Inspired by this human ability, a natural question arises: 
can LVLMs self-correct their output captions for better precisions and recalls? Some previous studies~\cite{huang2023large, qi2024moral} demonstrate that it is difficult for LLMs to make effective self-correction without extra training. 
In image captioning, we observe that directly prompting models to make self-corrections fails to get ideal results since it may remove correct descriptions and add hallucinations.

In order to teach LVLMs to make efficient self-correction, we propose SC-Captioner, a reinforcement learning framework that trains the model to remove hallucinations and add missing descriptions. Specifically, for each training image, the policy model first generates the initial caption given initial instruction. Then the first-turn instruction and caption are concatenated together with self-correction instruction and input back for self-corrected caption. After we get the paired captions, a correction-based reward is calculated. The formulation of the reward function is the core of self-correct training.

Given the predicted and reference captions, a scene-graph parsing algorithm is first employed to parse objects, attributes and relations from captions. Set difference is calculated between initial and self-corrected sets of elements to get the added and removed elements. Then these elements are matched against the reference sets to calculate correctness bonuses and mistake punishments. The former encourages accurate additions and removals, and the latter prevents incorrect insertions and deletions.

For more reliable and robust image caption evaluations, we propose a new reference-based evaluation metric. It makes improvements to some critical issues in the previous metric CAPTURE~\cite{dong2024benchmarking} like incomplete precision calculation, globally matched attributes and inefficient relation matching. Firstly, extra elements are introduced to prevent omissions in reference captions. Secondly, attributes are matched considering the objects they are belonging to. Furthermore, we replace the matching-based relation evaluation with a question-answer-based one. These refinements improve the rationality of evaluation.

We also construct a new dataset named RefinedCaps for training. This dataset comprises 6.5K images sampled from the COCO~\cite{chen2015microsoft} training set. Initial captions were generated using GPT-4~\cite{achiam2023gpt4}. These captions were then refined by human experts to correct any hallucinations and add omitted details to get the final captions. 
For testset, we sampled 500 images and their corresponding captions from DOCCI~\cite{onoe2024docci} and Localized-narratives~\cite{pont2020connecting}, respectively, forming two test subsets named DOCCI500 and COCO-LR500. On these subsets, we evaluated the performance of LVLM image captioners using a variety of metrics. 
Experiments are conducted and the results trained with RefinedCaps are presented in~\cref{sec:exp/main} and illustrated in~\cref{fig:1}. Simply conducting supervised fine-tuning on RefinedCaps can largely improve the caption performance. Direct Preference Optimization (DPO)~\cite{rafailov2023dpo} is first used to teach models to self-correct but fails to get higher results. As a comparison, our proposed method can consistently improve the image captioning results across different metrics. To further validate the robustness and generalization, we also conducted experiments using DOCCI as the training dataset. The results, detailed in the supplementary material, demonstrate that our method performs effectively on DOCCI, confirming its strong adaptability and reliability across different datasets.

Our contributions can be summarized as follows:

\begin{enumerate}
\item 
We are the first to introduce a self-correction training strategy into LVLM models on the image captioning task. 
This training strategy is a policy-gradient-based multi-turn reinforcement learning, where we design a novel correction-based reward function including correctness bonus and mistake punishment.

\item 
For image caption quality assessment, we propose a set of metrics refined from CAPTURE~\cite{dong2024benchmarking} that alleviate its problems of incomplete precision calculation, globally matched attributes and inefficient relation matching.

\end{enumerate}


\section{Related Works}

\subsection{Image Caption Generation}
Image caption generation is the process of generating textual descriptions of the images
by using natural language processing and computer vision. Previous approaches solve this task mainly by encoder-decoder paradigm with recursive, memory or attention mechanism~\citep{Vinyals_2015_CVPR, liu2017improved, cornia2020meshed, liu2017attention, huang2019attention, wang2022git}. Recently, more and more works~\citep{li2022blip, li2023blip2, fang2024vila, li2024monkey, wang2023caption, lu2025omnicaptioner} tend to use Large Vision-Language Models (LVLM) to help build more robust and comprehensive image captioner with differently designed vision-language alignment techniques and data augmentation pipelines. Some recent open-source LVLMs~\cite{liu2024llava1.5, chen2024far, wang2023cogvlm, wang2024qwen2} treat image captioning as one of important downstream training and testing tasks and achieve satisfactory caption performance.

Training data is one of the most important factors in LLM-driven tasks, and this holds true for the image captioning domain as well. In the early years, datasets such as Flickr30k~\cite{young2014image},
COCO Caption~\cite{chen2015microsoft} are widely used for training and evaluating.
However, the text length and complexity of these datasets are not enough for more detailed image captioning.
Recently, ShareGPT4V~\citep{chen2023sharegpt4v} carefully prompts GPT4-V to get large amount of more detailed image description. DCI~\cite{Urbanek2024DCI} consists of 7.8K images with long caption most of which is aligned to submasks of the image. DOCCI~\citep{onoe2024docci} offers 15K individually shot images and the corresponding refined caption produced by human annotators. 
Some works~\cite{rasheed2024glamm, li2024monkey, hsieh2024graph} also build image captioner training upon model-assisted data construction pipelines. 

Evaluation metrics for image captioning can be roughly divided into reference-based and reference-free ones. Reference-free metrics~\cite{hessel2021clipscore, sarto2023pacs, sarto2025bridge, lee2024fleur} utilize pre-trained vision-language models to assess the similarity of evaluated caption and image pair. Traditional reference-based metrics~\cite{papineni2002bleu, vedantam2015cider, banerjee2005meteor} compute n-gram matching score with ground truth captions. However, previous methods of both categories are not suitable for detailed caption evaluation due to limited training and weak generalization ability.
DC-Score~\cite{ye2025dcscore} introduces a framework to evaluate caption similarity at the level of primitive information units using GPT-4o, demonstrating strong correlation with human assessments. 
Some works~\cite{lee2021qace, liu2024playground3} propose to evaluate captions with questions. Although this approach is more reliable, it lacks completeness since elements covered in questions account for only a small fraction of all elements. 
A recently proposed metric CAPTURE~\cite{dong2024benchmarking} propose to solve the above problems by calculating F1 score of objects, attributes and relations parsed out from GT and candidate captions. The metrics proposed in this paper follows the approach of comparing these three components and further refines the irrationality of this method.

\subsection{Self-correction for Large Models}
Self-correction is an important capability of LLMs to refine their initial output given specific instructions during inference time~\cite{pan2024automatically, kamoi2024can}. 
Generally, self-correction approaches can be mainly divided into two primary types: intrinsic and extrinsic. Extrinsic self-correction~\cite{madaan2024self, yuan2025dolphin, team2025novelseek} relies on external feedback, such as input from humans or other evaluators, to explicitly identify problems in initial outputs and provide extra information for better response. While intrinsic ones~\cite{ganguli2023capacity} rely only on the intrinsic knowledge of the model and some extra instruction prompts to improve the answer quality. 

Some previous studies~\cite{huang2023large, qi2024moral} demonstrate that intrinsic self-correction is difficult to improve performance, but appropriate training can activate this ability. Several works have explored how to train models to perform self-correction effectively.  \cite{havrilla2024glore, welleck2022generating} rely on separate refinement model and fine-tuning to get better self-correction. \cite{qu2024recursive} forms multi-turn training as a Markov decision process and utilizes an iterative approach to collect data from stronger models and train the target model. \cite{kumar2024score} proposes a multi-turn reinforcement learning-based strategy to train LLMs and improves self-correction ability on math reasoning.

Previous self-correction training works mainly focus on pure LLM and reasoning tasks like math and coding. Few works have investigated self-correction in multimodal LLMs~\cite{he2024self, he2024topic} and we are the first to study self-correction in image captioning as far as we know.

\begin{figure*}[t]
  \centering
    \includegraphics[width=0.99\linewidth]{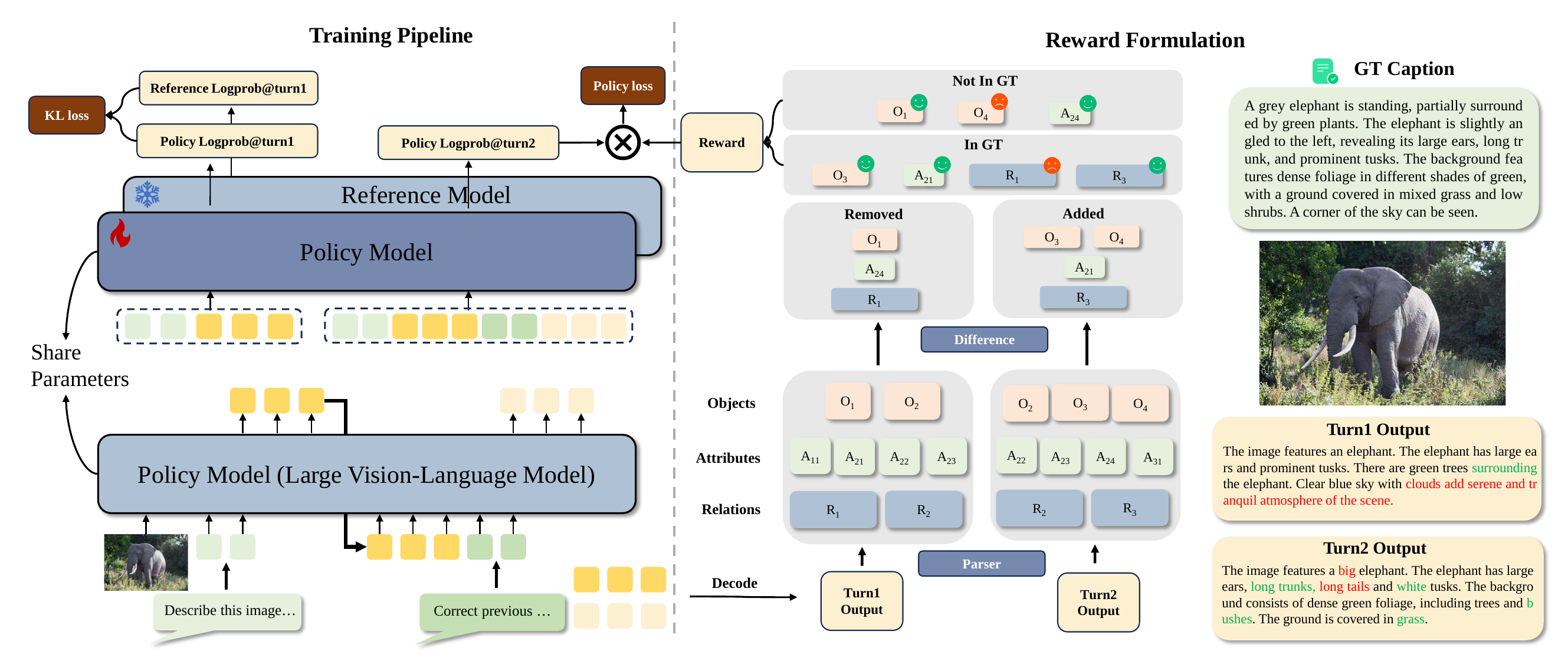}
    \caption{The overall framework of our proposed method. The training pipeline is shown on the left. It utilizes policy-gradient reinforcement learning to optimize the objective in~\cref{eq:objective}. On the right is the formulation of our correction-based reward function. Parsed sets of elements are subtracted and matched to calculate the final reward.}
    \label{fig:2}
\end{figure*}
\section{Method}

\subsection{Preliminaries}


Given a LVLM captioner parameterized by $\theta$, it will output initial caption $y_1$ for a given image-instruction pair $[I, x_1]$. Specifically, the caption $y_1$ is sampled from the policy disrtibution $\pi_\theta(\cdot|[I, x_1])$ generated by the model. For intrinsic self-correction, we offer a second-turn instruction $x_2$ that prompts the model to refine the initial caption $y_1$. Then the second-turn caption $y_2$ can be sampled from the policy distribution $\pi_\theta(\cdot|[I, x_1, y_1, x_2])$. With the high quality caption $y^*$ given, it can be regarded as ground-truth. The goal of good image captioners is to bring the contents of the generated captions $y_1$ and $y_2$ closer to $y^*$, regardless of the writing style.


As stated in prior works~\cite{rafailov2024direct, kumar2024score}, typical reinforcement learning objective is formulated as follows:

\begin{small}
\begin{equation}
    \max_{\theta} \mathbb{E}_{x \sim D, y \sim \pi_{\theta}(y|x)} \left[ r(x, y) - \beta \mathbb{D}_\text{KL} \left( \pi_{\theta}(y|x) \parallel \pi_\text{ref}(y|x) \right) \right], \\
    \label{eq:objective}
\end{equation}
\end{small}

\noindent where $r(x,y)$ is the reward of certain answer, and $\pi_\text{ref}$ denotes the policy of reference model (often the state of model at the beginning of training). We optimize this objective using simple policy-gradient methods and shape the reward directly using the proposed reward function rather than using a pre-trained reward model or training one from scratch.

\subsection{Multi-turn Reinforcement Learning with Policy-gradient}

The training pipeline of our proposed method is illustrated in the middle and right part of~\cref{fig:2}. Before training, the policy model and reference model are first initialized with the same checkpoint respectively. Only the policy model is updated during training and the reference model is frozen. Following the policy-gradient practice, the ``action" is first taken by generating the first-turn and second-turn captions. We first sample the initial caption $y_1$ from $\pi_\theta(\cdot|[I, x_1])$ with a high top-p value. Then the generated $y_1$ is then sent to the model to get the self-corrected caption $y_2$ sampled from $\pi_\theta(\cdot|[I, x_1, y_1, x_2])$. Once the two captions are generated, their corresponding probabilities --- $\pi_\theta(y_1|[I, x_1])$,  $\pi_\text{ref}(y_1|[I, x_1])$ and $\pi_\theta(y_2|[I, x_1, y_1, x_2])$ --- can each be computed with a single additional forward pass through the model. With these outputs and logits, the training objective can be formulated as follows:

\begin{small}
\begin{equation}
\begin{aligned}
    &L_\text{KL} = \text{log}(\pi_\theta(y_1|[I, x_1]))-\text{log}(\pi_\text{ref}(y_1|[I, x_1])) \\
    &L =  -R(y_1,y_2,y^*) \cdot \text{log}\pi_\theta(y_2|[I, x_1, y_1, x_2]) + \beta L_\text{KL}
\end{aligned}
\end{equation}
\end{small}

\noindent where $y^*$ is the ground truth (GT) caption, $R$ is the reward function that produces correction-based reward. The training objective can be divided into two parts. The first component is the policy loss aimed at enhancing the model's self-correction ability. Specifically, when the reward is positive, model parameters $\theta$ are updated in the direction of increasing the probability $\pi_\theta(y_2|[I, x_1, y_1, x_2])$; conversely, when the reward is negative, the parameters are adjusted to decrease this probability. The second component is the KL loss designed to maintain the model's capability to generate the initial caption. Together, these components ensure that the model not only generates plausible initial captions but also effectively refines them through self-correction.

\subsection{Reward Formulation}

Reward function is crucial for effective policy-gradient training, and it plays a vital role in enhancing the model's self-correction capabilities. The reward can be simply formulated as differences of certain evaluation metrics, such as BLEU~\cite{papineni2002bleu}, METEOR~\cite{banerjee2005meteor} or CAPTURE~\cite{dong2024benchmarking}. However, traditional metrics based on n-gram matching are sensitive to writing style, and CAPTURE is highly complex, weakening the correlation between metric changes and positive or negative correction. We propose a new reward function including correctness bonus and mistake punishment that directly reflect the effectiveness of self-correction made by model. To better disassemble the text changes, we employ three important concepts borrowed from scene graph: objects, attributes and relations. Attributes are descriptive properties of objects, such as color and material. Relations are connections between objects describing spatial or interaction situations.
The reward formulation is illustrated on the right side of~\cref{fig:2} and the two components is articulated as follows.

\textbf{Correctness bonus.} 
In short, correctness bonus is ``to reward the model for making correct additions and removals". Generally, it encourages the model to add descriptions not exists in initial caption but exists in GT caption and remove descriptions present in initial caption but absent in GT caption. 
Given model generated initial caption $y_1$, self-corrected caption $y_2$ and the GT caption $y^*$, the SOTA scene graph parser FACTUAL~\cite{li2023factual} is employed to extract objects, attributes and relations of them. Then rewards are calculated based on these three aspects respectively. Taking objects as an example, the parsed objects set is denoted as $\mathbf{O}_{y_1}, \mathbf{O}_{y_2}$ and $\mathbf{O}_{y^*}$. We calculate the maximum similarity sets $\mathbf{S_a}$ for objects added, and $\mathbf{S_r}$ for objects removed through self-correction:

\begin{small}
\begin{equation}
\begin{aligned}
    \mathbf{S_a}=\left\{ \max_{o^* \in \mathbf{O}_{y^*}} s(o_a, o^*) \mid o_a \in \mathbf{O}_{y_2} \setminus \mathbf{O}_{y_1} \right\}, \\
    \mathbf{S_r}=\left\{ \max_{o^* \in \mathbf{O}_{y^*}} s(o_r, o^*) \mid o_r \in \mathbf{O}_{y_1} \setminus \mathbf{O}_{y_2} \right\},
\end{aligned}    
\end{equation}
\end{small}
\noindent where $s(o, o^*)$ denotes the cosine similarity between the feature vectors of object phrase $o$ and GT object $o^*$, which are extracted using the Sentence Transformer model.

Soft scores $\sum_{s \in \mathbf{S_a}} (s - \tau_a)$ and $\sum_{s \in \mathbf{S_r}} (\tau_r - s)$ are used to measure the effectiveness of self-correction in adding descriptions of missing objects and removing descriptions of incorrect objects. We also employ hard scores $\sum_{s \in \mathbf{S_a}} \mathbf{1}_{\{s > \tau^{\prime}_a\}}$ and $\sum_{s \in \mathbf{S_r}} \mathbf{1}_{\{s < \tau^{\prime}_r\}}$ to award correct additions and removals. For attributes, only additions and removals of similar objects are computed. For relations, the triplets are concatenated as a whole to calculate the score.

\begin{figure*}[t]
  \centering
    \includegraphics[width=\linewidth]{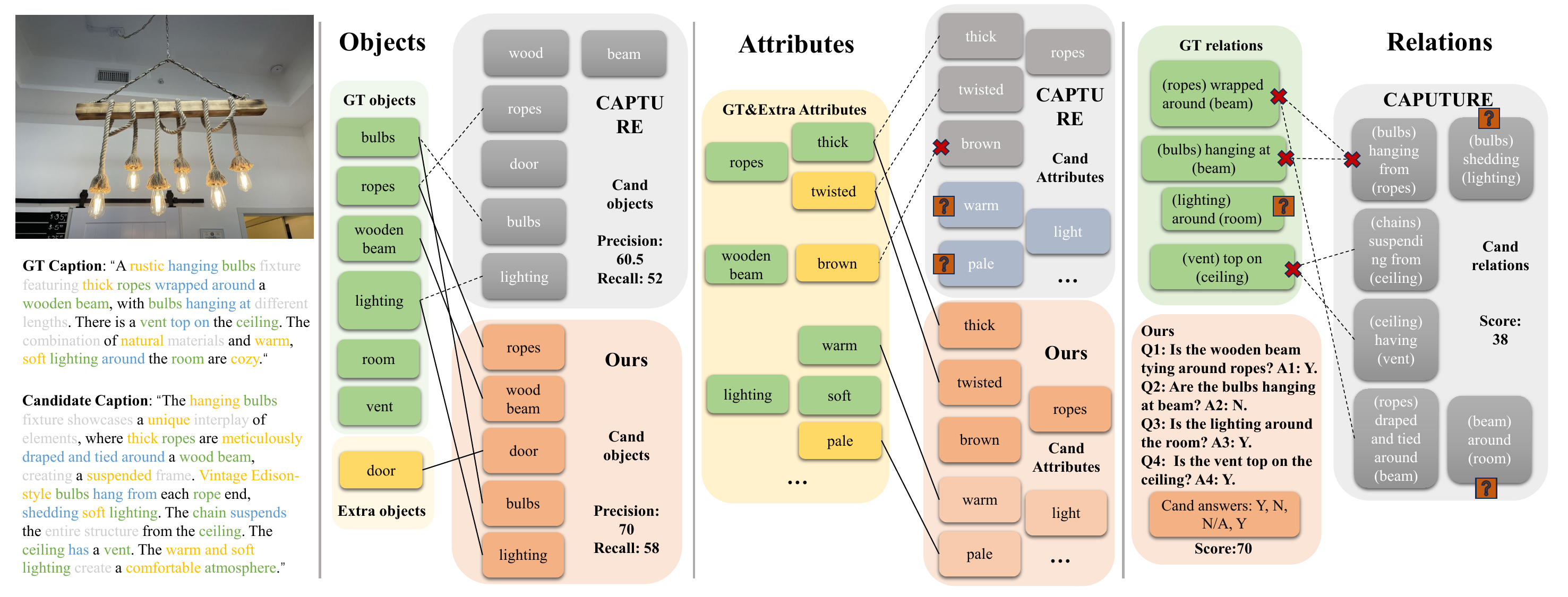}
    \caption{Visual comparison of CAPTURE metrics and our proposed metrics. Problems in CAPTURE such as incomplete precision calculation, globally matched attributes and inefficient relations matching are alleviated in the refined metrics.}
    \label{fig:3}
\end{figure*}

\textbf{Mistake punishment.}
Besides awarding correct additions and removals, this part of function lowers the reward when model making mistakes by adding hallucinations and removing correct descriptions. 
Generally, elements in $y_2$ but not in $y_1 \cup y^*$ and elements not in $y_2$ but in $y_1 \cup y^*$ will lower the final reward. Such punishment is not applied for relations to avoid confusion, since the match of relations is relatively complex.
Although in some cases the above reward formulation may punish added descriptions that exists in the image but not mentioned in $y^*$. Through our observation, this phenomenon is less common when the GT caption is relatively complete and does little harm to the results.
\section{Dataset and Metrics}

To facilitate better training and evaluating of the proposed method, GT captions with higher quality and more reliable evaluation metrics are essential. In this section, we will introduce the construction steps of our proposed training set, RefinedCaps, and the details of our proposed evaluation metrics.
\subsection{Dataset Construction}
For more reliable self-correction training, we build a new training dataset consists of 6.5K image-ground-truth-caption pairs with human assistance. The building process is stated as follows and illustrated in the left side of~\cref{fig:2}: 

1). 6.5K images are randomly selected from the COCO 2017 trainset.

2). GPT-4o is employed to generate initial caption for these images. We also prompts GPT-4o to remove sentences and phrases describing the atmosphere like \textit{``adding to xxx/creating xxx scene"} from the initial caption in a different dialogue.

3). The initial captions are prone to be incomplete and have hallucinations. To improve the precision of the captions, we invite human experts to remove or revise all the hallucinations in the initial caption. Hallucinations include but are not limited to wrong/non-existing objects/attributes/relations, wrong counting and positions, ambiguous descriptions.

4). To enhance the recall of captions, human experts are asked to add descriptions of objects and important relations between objects in the image that were not included in the initial caption. At least 80\% of the objects in the image should appear in the caption. Also, for main objects in the foregrounds, it is ensured that there is at least one attribute description word or phrase.

\noindent The newly-constructed dataset is named \textbf{RefinedCaps}. By incorporating human expert revisions, RefinedCaps ensures that the captions are precise and cover a significant portion of the objects and their relationships within the images.

\subsection{Evaluating Metrics}
\label{sec:dataset/benchmark}


Following CAPTURE~\cite{dong2024benchmarking}, the whole evaluation is divided into three parts: objects, attributes and relations. The FACTUAL scene graph parser is employed to extract objects, attributes and relations in the beginning. However, the CAPTURE metric has some problems like misalignment of certain forms of phrases, incomplete precision evaluation and inefficient relation matching. ~\Cref{fig:3} shows some examples of imperfections of CAPTURE metric and our improvements stated below. 

\begin{table*}[t]
  \centering
  \resizebox{0.99\linewidth}{!}
  {
  \begin{tabular}{lccccccccccc}
    \toprule
    \multirow{2}{*}{Base Model} & \multirow{2}{*}{Post-training} & \multirow{2}{*}{BLEU-4} & \multirow{2}{*}{METEOR} & \multirow{2}{*}{CAPTURE} & \multicolumn{3}{c}{Objects} & \multicolumn{3}{c}{Attributes} & Relations \\
    \cmidrule(lr){6-8} \cmidrule(lr){9-11} \cmidrule(lr){12-12}
    
    & & & & & Precision & Recall & F1 & Precision & Recall & F1 & QA \\
    \midrule


    \multirow{8}{*}{LLaVA-1.5-7B} & Zero-shot & 24.36 & 14.59 & 50.72 & 81.13 & 48.37 & 59.41 & 64.12 & 39.78 & 48.01 & 9.19 \\
    & Zero-shot$^*$& 23.09 & 14.08 & 50.47 & \bf{81.40} & 48.25 & 59.25 & 64.70 & 40.10 & 48.39 & 9.23 \\
    & SFT& 40.08 & 21.25 & 60.92 & 78.15 & 62.20 & 68.44 & 66.85 & 46.27 & 53.93 & 19.87 \\
    & SFT$^*$& 40.14 & 21.33 & 61.03 & 78.15 & 62.39 & 68.55 & 67.05 & 46.36 & 54.05 & 20.11 \\
    & SFT+DPO& 42.38 & 22.90 & 61.30 & 76.64 & 62.88 & 68.26 & 66.22 & 46.83 & 54.15 & 20.64 \\
    & SFT+DPO$^*$& 42.67 & 23.47 & 61.09 & 75.31 & 64.75 & 68.89 & 65.64 & 46.29 & 53.61 & 21.73 \\
    & SFT+Ours& 42.60 & 22.84 & 62.20 & 77.92 & 64.08 & 69.61 & \bf{67.81} & 48.72 & 55.94 & 21.93 \\
    & SFT+Ours$^*$& \bf{43.04} & \bf{23.88} & \bf{62.29} & 77.58 & \bf{66.05} & \bf{70.30} & 67.64 & \bf{50.70} & \bf{57.10} & \bf{22.69} \\

    \midrule

    \multirow{8}{*}{Qwen2-VL-7B} & Zero-shot & 29.39 & 16.59 & 57.96 & \bf{83.69} & 56.79 & 66.47 & 69.96 & 43.27 & 52.65 & 17.57 \\
    & Zero-shot$^*$& 27.16 & 15.88 & 57.62 & 83.61 & 56.10 & 65.88 & \bf{70.36} & 43.49 & 52.92 & 17.01 \\
    & SFT& 40.92 & 22.04 & 62.05 & 79.99 & 62.68 & 69.50 & 68.85 & 47.57 & 55.50 & 27.65 \\
    & SFT$^*$& 38.93 & 21.34 & 62.02 & 80.46 & 62.19 & 69.32 & 69.18 & 47.91 & 55.78 & 27.93 \\
    & SFT+DPO& 43.13 & 22.88 & 62.56 & 79.86 & 64.85 & 70.77 & 67.74 & 47.92 & 55.36 & 26.44 \\
    & SFT+DPO$^*$& 44.49 & 23.84 & 62.51 & 78.78 & 65.44 & 70.67 & 67.67 & 48.30 & 55.60 & 27.33 \\
    & SFT+Ours& 43.70 & 23.35 & 63.00 & 80.48 & 64.48 & 70.74 & 69.68 & 48.83 & 56.61 & 29.06 \\
    & SFT+Ours$^*$& \bf{44.88} & \bf{25.18} & \bf{63.34} & 80.20 & \bf{66.00} & \bf{71.63} & 69.54 & \bf{50.34} & \bf{57.67} & \bf{30.51} \\

    \bottomrule
  \end{tabular}
  }
  \caption{Results on DOCCI500, the overall score of BLEU-4, METEOR, CAPTURE and seven aspects of our proposed evaluation metrics is reported. $^*$ denotes metrics of the self-corrected captions. Best results are highlighted in bold.}
  \label{tab:main1}
\end{table*}

\textbf{Objects.} For objects, recall means the ratio of objects in the GT caption appears in the candidate caption. And a high precision represents less hallucination objects that appears in candidate caption but not in the image. In precision computing, CAPTURE only match candidate objects with objects parsed from GT captions. This approach may fail to match objects that appears in the image but not mentioned in the GT caption. To tackle this, extra objects generated by GPT-4o and evaluated by human experts are merged to create wider sets of objects for precision computing. 

\textbf{Attributes.} In CAPTURE, the process of calculating attribute scores is directly matching all attributes, regardless of the corresponding objects. This may cause mismatch between similar attributes belonging to different objects. The approach has been modified to only match attributes for the same or similar objects. Similar to objects, GPT-4o generated attributes for all objects are adopted in precision computation. For each sample, candidate attributes set $\{(o^c_i, a^c_{i1}, a^c_{i2}, \ldots, a^c_{in_i}) | i=1,2,\ldots,N_c\}$, GT attributes set $\{(o^g_i, a^g_{i1}, a^g_{i2}, \ldots, a^g_{in_i}) | i=1,2,\ldots,N_g\}$ and GPT-4o expanded GT attributes set $\{(o^e_i, a^e_{i1}, a^e_{i2}, \ldots, a^e_{in_i}) | i=1,2,\ldots,N_e\}$, the precision and recall of attributes can be calculated as follows:

\begin{small}
\begin{equation}
\begin{aligned}
  &P = \frac{\sum\limits_{i} \left[s(o_i^c,o_j^e)\cdot \sum\limits_{m} \max\limits_{n} s(a^c_{im}, a^e_{jn})\right]}{\sum\limits_{i} \left[s(o_i^c,o_j^e) \cdot n_i\right]}, j=\argmax_n\left(s(o_i^c,o_n^e)\right) \\
  &R = \frac{\sum\limits_{i} \left[s(o_i^g,o_j^c)\cdot \sum\limits_{m} \max\limits_{n} s(a^g_{im}, a^c_{jn})\right]}{\sum\limits_{i} \left[s(o_i^g,o_j^c) \cdot n_i\right]}, j=\argmax_n\left(s(o_i^g,o_n^c)\right)
  \label{eq:important}
\end{aligned}
\end{equation}
\end{small}

\noindent where $s(a,b)=f(a)\cdot f(b)$ is the semantic similarity of two words, and $f$ is a pre-trained text encoder model.

\begin{table*}[t]
  \centering
  \resizebox{0.99\linewidth}{!}
  {
  \begin{tabular}{lccccccccccc}
    \toprule
    \multirow{2}{*}{Base Model} & \multirow{2}{*}{Post-training} & \multirow{2}{*}{BLEU-4} & \multirow{2}{*}{METEOR} & \multirow{2}{*}{CAPTURE} & \multicolumn{3}{c}{Objects} & \multicolumn{3}{c}{Attributes} & Relations \\
    \cmidrule(lr){6-8} \cmidrule(lr){9-11} \cmidrule(lr){12-12}
    
    & & & & & Precision & Recall & F1 & Precision & Recall & F1 & QA \\
    \midrule


    \multirow{8}{*}{LLaVA-1.5-7B} & Zero-shot & \bf{38.48} & 20.60 & 44.75 & \bf{81.20} & 59.02 & 67.56 & 58.61 & 37.20 & 42.34 & 14.38  \\
    & Zero-shot$^*$ & 37.90 & 20.17 & 44.89 & 80.86 & 59.30 & 67.57 & 60.80 & 38.63 & 43.85 & 14.95 \\
    & SFT& 33.81 & 26.29 & 46.62 & 77.09 & 71.16 & 73.45 & 66.65 & 50.57 & 54.25 & 28.59 \\
    & SFT$^*$& 33.60 & 26.37 & 46.61 & 77.15 & 71.55 & 73.72 & \bf{66.78} & 50.70 & 54.43 & 27.82 \\
    & SFT+DPO& 31.05 & 26.58 & 46.42 & 76.01 & 73.10 & 73.95 & 65.28 & 52.32 & 54.79 & 28.88 \\
    & SFT+DPO$^*$& 28.84 & 26.81 & 45.70 & 74.78 & 74.18 & 73.98 & 64.39 & 52.04 & 54.24 & 29.85 \\
    & SFT+Ours& 32.72 & 26.76 & 46.57 & 76.89 & 73.49 & 74.64 & 66.18 & 51.90 & 54.87 & 32.37 \\
    & SFT+Ours$^*$& 31.60 & \bf{27.05} & \bf{47.11} & 76.43 & \bf{75.65} & \bf{75.20} & 65.79 & \bf{53.01} & \bf{55.35} & \bf{33.63} \\

    \midrule

    \multirow{8}{*}{Qwen2-VL-7B} & Zero-shot & \bf{39.57} & 20.42 & 46.52 & 81.12 & 61.82 & 69.47 & 66.48 & 42.86 & 48.68 & 20.47 \\
    & Zero-shot$^*$& 38.99 & 19.58 & 46.61 & \bf{81.42} & 61.71 & 69.52 & 66.78 & 42.97 & 48.81 & 21.16 \\
    & SFT& 34.59 & 26.49 & 47.14 & 78.64 & 73.24 & 75.37 & 69.01 & 53.15 & 56.54 & 36.39 \\
    & SFT$^*$& 34.29 & 26.57 & 47.08 & 78.76 & 73.27 & 75.43 & \bf{69.02} & 53.09 & 56.57 & 36.88 \\
    & SFT+DPO& 31.78 & 26.88 & 46.71 & 77.51 & 73.38 & 74.86 & 67.68 & 52.28 & 55.35 & 36.23\\
    & SFT+DPO$^*$& 30.34 & 26.77 & 46.44 & 77.01 & 74.23 & 75.14 & 67.28 & 52.65 & 55.79 & 37.21\\
    & SFT+Ours& 35.29 & 27.00 & 47.28 & 78.83 & 74.34 & 76.07 & 68.66 & 54.44 & 57.16 & 37.73 \\
    & SFT+Ours$^*$& 35.05 & \bf{27.34} & \bf{47.51} & 78.72 & \bf{75.01} & \bf{76.37} & 68.43 & \bf{55.11} & \bf{57.56} & \bf{38.51}\\

    \bottomrule
  \end{tabular}
  }
  \caption{Results on COCO-LN500, the overall score of BLEU-4, METEOR, CAPTURE and seven aspects of our proposed evaluation metrics is reported. $^*$ denotes metrics of the self-corrected captions. Best results are highlighted in bold.}
  \label{tab:main2}
\end{table*}

\textbf{Relations.} The situation of relations is much more complex than objects and attributes. Simply matching the concatenated triplets of two objects and one conjunction is not an ideal solution of relation evaluation. Also, results reported by CAPTURE show that the whole evaluation consistency is not significantly affected even if relation scores are removed. Inspired by question-based metrics, we propose to evaluate relations with questions. For each testing image, five questions about relations and positions of objects with corresponding correct answer are raised according to both the image and GT caption using GPT-4o and checked by human experts. In the evaluation stage, open-source language models are prompted to answer these questions based on the candidate caption for each image. The total accuracy of all questions are computed as the metric for relations evaluation.

\section{Experiments}

\subsection{Experimental Settings}

To effectively evaluate the captioning performances, two human-annotated datasets from different sources are selected as test sets. Firstly, we random sample 500 images and the corresponding captions from the DOCCI~\cite{onoe2024docci} test set. This dataset is referred to as \textbf{DOCCI500}. Secondly, other 500 images are selected from the Localized-narratives~\cite{pont2020connecting} test set in COCO2017. We have made a filter and ensure each caption contains human-related words and contains at least 60 words, aiming to complement DOCCI500, which is largely devoid of human-centric descriptions. This dataset is referred to as \textbf{COCO-LN500}. Various metrics including BLEU-4, METEOR, CAPTURE and our proposed metrics are used for evaluation.

Experiments are built upon two baseline LVLMs: LLaVA-1.5-7B~\cite{liu2024llava1.5} and Qwen2-VL-7B~\cite{wang2024qwen2}. These two LVLMs already have strong abilities to generate detailed image caption. In this section, evaluations based on our sampled datasets DOCCI500 and COCO-LN500 are first conducted for these two baselines. Then the collected RefinedCaps dataset with 6.5K image-ground-truth-caption pairs are used to further boost the performance of them. Supervised fine-tuning (SFT) on RefinedCaps is first conducted to align the model output distribution to that of the training set for further post-training. We implement another method Direct Preference Optimization (DPO) for self-correction training and take its performance into comparison. Specifically, Initial and corrected captions $y_1^*$ and $y_2^*$ are first sampled for all training images. The input of DPO training is $[I, x_1, y_1^*, x_2]$, the ``rejected" text is $y_2^*$ and the ``chosen" text is the corresponding GT caption $y^*$.
For experimental purpose and fair comparison, no prompt engineering attempts are made. The instructions $x_1$ and $x_2$ remains simple and the same among all experiments. Simple LoRA~\cite{hu2022lora} fine-tuning is adopted for all training processes.
All the implementations are built based on the LLaMA-Factory~\cite{zheng2024llamafactory} codebase and run on 8$\times$A800 GPUs.

\subsection{Experimental results}
\label{sec:exp/main}
The main experimental results are shown in~\cref{tab:main1} and~\cref{tab:main2}. In~\cref{tab:main1}, models are tested on the DOCCI500 dataset which has more detailed GT captions and contains few human figures. While in~\cref{tab:main2}, the COCO-LN500 dataset including images mostly featuring humans and compact captions in narrative style is used as testing set. Both initial and self-corrected captions are evaluated. 

As can be seen in the tables, supervised fine-tuning with the proposed RefinedCaps dataset can significantly improve the performance of the baseline model. However, for baseline and SFT models, the metrics of self-corrected captions are not higher than those of initial captions, indicating that the self-correcting capability of baseline and SFT models is not enough. DPO teaches the models to prefer correcting the initial captions to the ground-truth captions rather than to their original self-correction results. Metrics in the tables show that models trained with DPO has high Recall scores but low Precision scores. 
The results demonstrate that training by DPO can indeed enhance the self-correction ability of models to some extent. However, DPO seems can only teach the models to make self-correct action but fails to teach them how to correctly remove hallucination contents and add missing contents. 
Contrastively, our proposed method can consistently improve the CAPTURE score, the recall of objects and attributes, and accuracy of relation QA, while maintaining similar precision levels as SFT. 
Moreover, results using DOCCI as training dataset reported in the supplementary material show similar trends, demonstrating the robustness and versatility of the proposed method. 

\begin{table*}[t]
  \centering
  \resizebox{0.99\linewidth}{!}
  {
  \begin{tabular}{ccccc|cccccccc}
    \toprule
    \multicolumn{5}{c|}{Reward Formulation} & \multirow{2}{*}{CAPTURE} & \multicolumn{3}{c}{Objects} & \multicolumn{3}{c}{Attributes} & Relations \\
    \cmidrule{1-5}  \cmidrule(lr){7-9} \cmidrule(lr){10-12} \cmidrule(lr){13-13} 
    
     CAPTURE & Object & Attribute & Relation & Mistake & & Precision & Recall & F1 & Precision & Recall & F1 & QA \\
    \midrule

       &&&&& 62.02 & 80.46 & 62.19 & 69.32 & 69.18 & 47.91 & 55.78 & 27.93 \\

     \checkmark&&&& & 62.34 & 79.76 & 64.37 & 70.43 & 68.94 & 49.50 & 57.02 & 29.66 \\
     &\checkmark&&& \checkmark& 62.83 & 79.64 & 66.35 & 71.44 & 69.02 & 48.69 & 56.30 & 29.42\\
     &&\checkmark&& \checkmark& 62.40 & \bf{80.51} & 63.63 & 70.05 & 69.03 & 49.88 & 57.31 & 28.47 \\
     &&&\checkmark& \checkmark & 62.25 & 79.73 & 63.32 & 69.51 & 68.55 & 48.13 & 55.72 & 28.87 \\
     &\checkmark &\checkmark &\checkmark& & 63.20 & 77.52 & \bf{67.56} & 71.54 & 67.04 & \bf{50.96} & 57.11 & 29.96 \\
    \checkmark&\checkmark&\checkmark&\checkmark& \checkmark& 63.27 & 79.93 & 66.12 & 71.57 & 68.32 & 49.61 & 56.83 & 30.33 \\
    &\checkmark&\checkmark&\checkmark& \checkmark& \bf{63.34} & 80.20 & 66.00 & \bf{71.63} & \bf{69.54} & 50.34 & \bf{57.67} & \bf{30.51}\\


    \bottomrule
  \end{tabular}
  }
  \caption{Ablation study on components in reward formation in the reinforcement learning process. Effects of adopting CAPTURE score, match of objects/attributes/relations and mistake punishment in rewards are studied. Captions after self-correction are tested.}
  \label{tab:ablation_reward}
\end{table*}

Moreover, in the COCO-LN500 scenario, where the ground truth captions are generally shorter, the CAPTURE metric tends to output a lower precision score. This is because the shorter GT captions do not mention some objects and details present in the images, leading to a penalized precision score for correctly identifying these elements. The highest performance of baseline in BLEU-4 is also due to this.
Our proposed evaluation metrics, by incorporating additional information, avoids this penalty and maintains a more reasonable precision score.

\subsection{Ablation study}
\label{sec:exp/ablation}
To evaluate the effectiveness of designs in our proposed method, ablation studies on the designs of the reward function are conducted. We perform these ablations using Qwen2-VL-7B as base model and test the self-corrected caption performance on DOCCI500 using CAPTURE and our metrics.


The results of ablation studies are presented in~\cref{tab:ablation_reward}. Firstly, directly using CAPTURE score as the reward is tested. We formulate the reward function with reward shaping similar to~\cite{kumar2024score} as follows:

\begin{equation}
    R = C_2 + C_1 + \beta(C_2-C_1),
\end{equation}

\begin{figure}[t]
\vspace{-0.3cm}
  \centering
  \includegraphics[width=\linewidth]{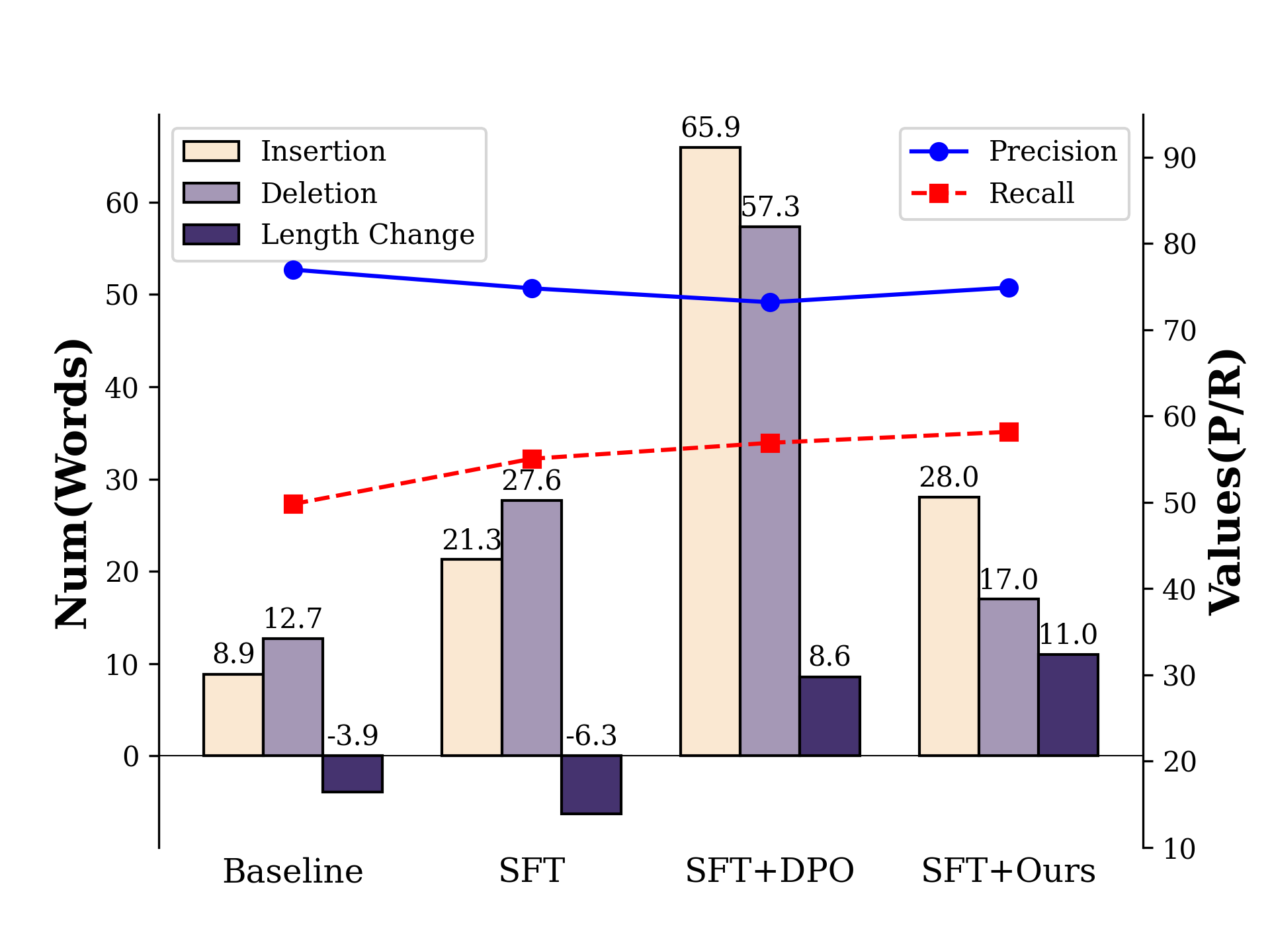}

   \caption{Comparisons of word number changes between baseline, SFT, DPO and our method. Precisions and Recalls of each models are also presented in the form of lines.}
   \label{fig:4}
\end{figure}

\noindent where $C_1$ denotes the CAPTURE score of initial caption and $C_2$ denotes that of the self-corrected caption. $\beta$ is the reward shaping function coefficient, aiming to encourage the model to produce different caption in second turn. This approach can improve the performance of captioning and self-correct, but is hard to get higher scores. It may because the total CAPTURE score is relatively complex and response weakly to both accurate and incorrect modifications. Also, CAPTURE metrics may suffer from the problems mentioned in~\cref{sec:dataset/benchmark} and~\cref{fig:3}. In contrast, the proposed reward function based on correctness bonus and mistake punishment can better guide the self-correction training. We then investigated each components of the proposed reward. As can be seen in~\cref{tab:ablation_reward}, rewards on objects, attributes and relations can notably improve the performance on these three concepts respectively. And they can combine to achieve better results. We also removed the mistake punishment from the reward function and found that the precision scores dropped significantly. While the mistake punishment may slightly reduce the recall score, it ensures the precision of the additions and removals made in self-correction.


\subsection{Statistical analysis}

To illustrate the self-correction action more clearly, we count the number of words inserted, deleted in the self-corrected caption compared to the initial caption, and the change of total length of captions. Results are shown in~\cref{fig:4}. Precisions and recalls are also presented in the figure. Firstly, SFT performs more insertions and deletions than baseline, and both of them reduce the length of caption after self-correction. The insertion and deletion of DPO far exceed other methods, and most of these changes are not effective for improving the quality of caption. Our method shows more insertions and less deletions than SFT, indicating that our method learns to properly insert and delete words to get better precision and recall.

\section{Conclusion}


In this work, we introduce SC-Captioner, a reinforcement learning framework that enhances the self-correction ability of image caption models. By designing a reward function that incentivizes accurate corrections and penalizes hallucinations, we significantly improve the quality of generated captions. We also refine the CAPTURE metrics to better evaluate detailed image captions and create the RefinedCaps dataset, comprising 6.5K diverse images from the COCO dataset. Experiments show that SC-Captioner outperforms existing methods, making substantial improvements in image captioning accuracy and reliability. 

\section{Acknowledgment}

This work is supported by National Key Research and Development Program of China (No. 2022ZD0160101), Shanghai Natural Science Foundation (No. 23ZR1402900), Shanghai Science and Technology Commission Explorer Program Project (24TS1401300), Shanghai Municipal Science and Technology Major Project (No.2021SHZDZX0103).
The computations in this research were performed using the CFFF platform of Fudan University.
{
    \small
    \bibliographystyle{ieeenat_fullname}
    \bibliography{main}
}

\clearpage
\setcounter{page}{1}
\maketitlesupplementary

\section{Prompt Templates}

We follow the official instruction of each LVLM and adopt simple prompt for image captioning and self-correction. For relation evaluation, we prompt open-source language models to answer the given questions based on the candidate captions. The Prompts used are illustrated as follows:

\begin{figure}[h]
  \centering
  \includegraphics[width=\linewidth]{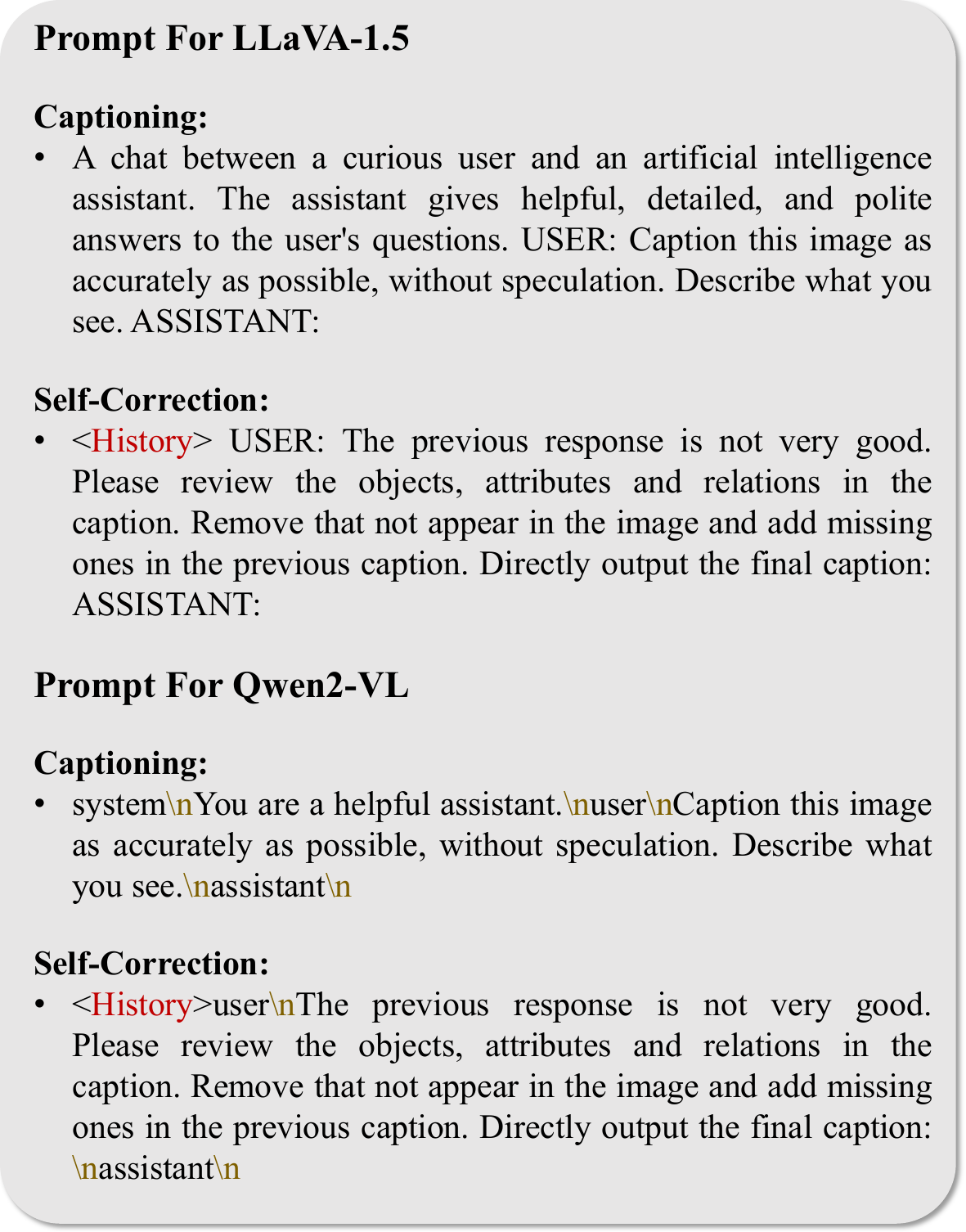}

   \caption{Prompts for image captioning and self-correction.}
   \label{fig:prompt1}
\end{figure}

\begin{figure}[h]
  \centering
  \includegraphics[width=\linewidth]{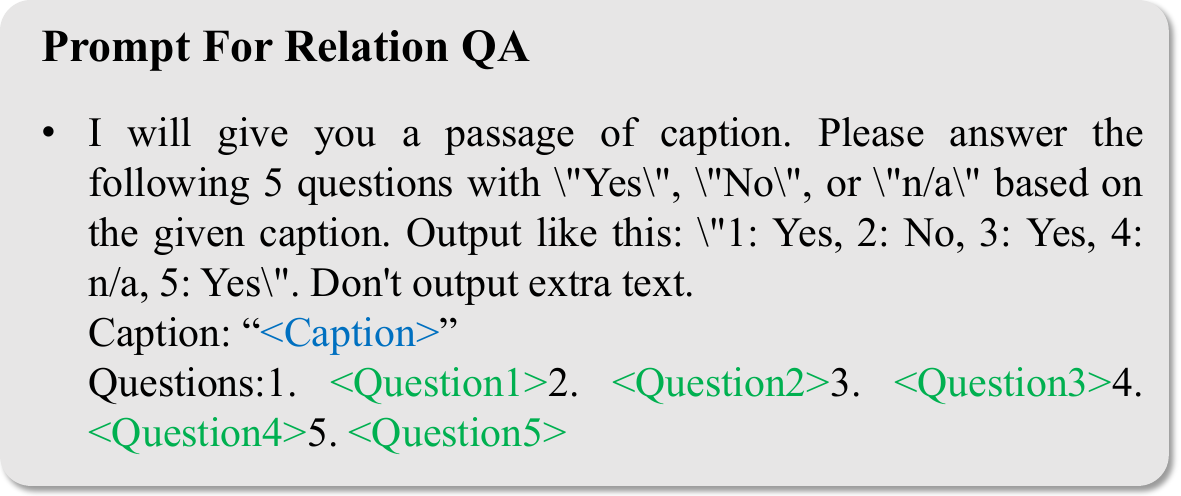}

   \caption{Prompts for Relation evaluation via QA.}
   \label{fig:prompt2}
\end{figure}

\section{Human Consistency of Proposed Metric}

We conducted an extra experiment to investigate how well our proposed metric aligns with human judgement. We randomly select 100 images in DOCCI500 and ask 4 human annotators to sort the captions provided by 4 different models, while considering both precision and recall. Then we calculate the Kendall's $\tau$ of BLEU-4, METEOR, CAPTURE and our metric (weighting 5,5,2 for objects, attributes and relations). The results in~\cref{tab:supp_humancorr} show that our metric has better alignment with human judgement.

\begin{table}[h]
  \centering
  \renewcommand{\arraystretch}{0.8}
  \resizebox{0.95\linewidth}{!}
  {
  
  \begin{tabular}{lcccc}
    \toprule
      & BLEU-4 & METEOR & CAPTURE & Our Metric  \\

    \midrule

     Kendall's $\tau$ & 30.73 & 32.26 & 37.66 & \bf{45.99} \\

    \bottomrule
  \end{tabular}
  }
  \caption{Correlation of metrics and human judgments. Our metric gets higher score than CAPTURE and traditional metrics.}
  \label{tab:supp_humancorr}
\end{table}

\section{Statistics of Captions}
We have made some analyses on different datasets including RefinedCaps, DOCCI, DCI and Localized Narratives in~\cref{tab:supp_stats}. As shown in the table, captions in our proposed dataset are relatively long and have more densely packed descriptions about objects, attributes and relations.
\begin{table}[h]
\vspace{-0.1cm}
  \centering
  \resizebox{0.99\linewidth}{!}
  {
  \begin{tabular}{lcccc}
    \toprule
     Dataset & Words & Objects & Attributes & Relations \\

    \midrule


    RefinedCaps & 120.53 & \bf{16.82} & \bf{16.14} & \bf{11.90} \\
    DOCCI & 121.91 & 13.33 & 14.29 & 10.50 \\
    DCI & \bf{133.23} & 15.90 & 14.14 & 10.90 \\
    Localized Narratives* & 40.47 & 6.89 & 1.52 & 4.45 \\
    COCO-LN500 & 77.46 & 11.26 & 2.89 & 7.68 \\

    \bottomrule
  \end{tabular}
  }
  \caption{Statistics across different datasets. * denotes that only a subset on COCO is selected. RefinedCaps has the highest element density.}
  \vspace{-0.2cm}
  \label{tab:supp_stats}
\end{table}

\section{Additional Experiments}
\label{sec:add_exp}

\subsection{Results of Using Public Dataset for Training}
We also use the training set of DOCCI~\cite{onoe2024docci} which consists of 9.7K image-caption pairs as the training set for supervised fine-tuning and self-correction training of Qwen2-VL. Metrics for both the initial and self-corrected captions are shown on both DOCCI500 and COCO-LN500 datasets in~\cref{tab:appendix_final}. It can be seen that our proposed method outperforms SFT and DPO by a considerable margin, especially in crucial F1 and QA metrics, demonstrating the universality of our method. 

\begin{table*}[t]
  \centering
  \resizebox{0.99\linewidth}{!}
  {
  \begin{tabular}{lccccccccccc}
    \toprule
    \multirow{2}{*}{Scenario} & \multirow{2}{*}{Post-training} & \multirow{2}{*}{BLEU-4} & \multirow{2}{*}{METEOR} & \multirow{2}{*}{CAPTURE} & \multicolumn{3}{c}{Objects} & \multicolumn{3}{c}{Attributes} & Relations \\
    \cmidrule(lr){6-8} \cmidrule(lr){9-11} \cmidrule(lr){12-12}
    
    & & & & & Precision & Recall & F1 & Precision & Recall & F1 & QA \\
    \midrule

    \multirow{6}{*}{Same-Domain} & SFT & 40.29 & 25.44 & 62.53 & 78.01 & 65.30 & 70.31 & 67.13 & 49.33 & 56.08 & 25.43 \\
    & SFT$^*$ & 41.78 & 26.13 & 62.72 & 77.70 & 65.87 & 70.52 & 66.99 & 49.94 & 56.47 & 25.96 \\
    & SFT+DPO & 42.79 & 25.90 & 63.04 &76.95 & 66.65 & 70.71 & 66.44 & 50.17 & 56.40  & 27.25 \\
    & SFT+DPO$^*$ & \bf{43.20} & \bf{26.80} & 63.28 & 75.53 & \bf{67.90} & 71.02 & 64.52 & \bf{51.34} & 56.46 & 27.82 \\
    & SFT+Ours & 41.10 & 26.11 & 63.47 & \bf{79.09} & 66.30 & 71.45 & \bf{70.09} & 50.04 & 57.63 & 26.68 \\
    & SFT+Ours$^*$ & 41.95 & 26.38 & \bf{63.83} & 78.85 & 67.59 & \bf{72.05} & 69.77 & 50.64 & \bf{58.00} & \bf{28.58} \\

    \midrule

    \multirow{6}{*}{Cross-Domain} & SFT & \bf{34.93} & 25.97 & 47.63 & 78.03 & 70.42 & 73.43 & 67.91 & 53.51 & 56.31 & 30.06 \\
    & SFT$^*$ & 33.22 & 26.03 & 47.40 & 77.70 & 71.49 & 73.86 & 67.42 & 53.47 & 56.09 & 30.58 \\
    & SFT+DPO & 32.34 & 25.81 & 46.73 & 76.22 & 71.09 & 72.94 & 66.87 & 53.96 & 56.26 & 31.01 \\
    & SFT+DPO$^*$ & 28.81 & 25.84 & 45.96 & 74.30 & \bf{73.05} & 73.04 & 65.00 & 52.78 & 54.79 & 30.84 \\
    & SFT+Ours & 32.69 & 25.83 & \bf{47.97} & \bf{78.64} & 70.94 & 74.13 & \bf{69.05} & 53.85 & \bf{57.00} & 30.75 \\
    & SFT+Ours$^*$ & 32.75 & \bf{26.17} & 47.92 & 78.58 & 72.09 & \bf{74.60} & 69.01 & \bf{54.03} & 56.93 & \bf{32.01}  \\


    \bottomrule
  \end{tabular}
  }
  \caption{Results of Qwen2-VL-7B training with DOCCI training set. BLEU-4, METEOR, CAPTURE and seven aspects of our proposed evaluation metrics are reported. $^*$ denotes metrics of the self-corrected captions. ``Same-Domain'' refers to the performance on DOCCI500 test set which has the same image and caption distribution as the training set. ``Cross-Domain'' denotes the performance on COCO-LN500 which has different distribution from the training set. Best results are highlighted in bold. Our proposed method outperforms baseline and DPO. Comparisons between ``Same-Domain'' in this table and~\cref{tab:main1} show that our proposed RefinedCaps dataset achieves comparable performance with the training set of the same domain (71.63 vs. 72.05 in Objects F1). However, the DOCCI training set performs worse in cross-domain scenario compared to the results in~\cref{tab:main2} (74.60 vs. 76.37 in Objects F1).}
  \label{tab:appendix_final}
\end{table*}

Since DOCCI500 test set is sampled from DOCCI, it can be referred to as a same-domain scenario. In contrast, COCO-LN500 represents a cross-domain scenario. The results in~\cref{tab:appendix_final} can be compared with those in~\cref{tab:main1} and ~\ref{tab:main2} to investigate the influence of training datasets. On DOCCI500 which is in the same domain as DOCCI training set, results of model trained on RefinedCaps are still comparable (71.63 vs. 72.05 in Objects F1, 57.67 vs. 58.00 in Attributes F1, 30.51 vs. 28.58 in Relations QA). However, on COCO-LN50 which is not the same domain as DOCCI training set, models trained on DOCCI performs much worse (74.60 vs. 76.37 in Objects F1, 56.93 vs. 57.56 in Attributes F1, 32.01 vs. 38.51 in Relations QA).
The above in-domain and cross-domain analyses demonstrate that the generalization and adaptation ability of our proposed RefinedCaps dataset is stronger than DOCCI dataset in terms of supervised fine-tuning and self-correction training.

\subsection{More Comparisons with Other Methods}

To further demonstrate the effectiveness of our proposed method, more comparative experiments are conducted with different methods and the results are shown in~\cref{tab:supp_ablation}. The first four lines are results reported in~\cref{tab:main1}. DiscriTune is a reinforcement learning method introduced in~\cite{dessi2023cross}, which utilizes CLIP~\cite{radford2021learning} to produce reinforcement learning loss. Line 5 shows that this method fails to achieve satisfactory results. It may because CLIP truggles to effectively distinguish differences when dealing with very long captions. We also try to calculate the reward solely from the output of the first turn and put the results in line 6. It can boost performance, but fails to exceed the proposed two-step approach, demonstrating the necessity of self-correction. Additionally, we tested an extra baseline one where the SFT model is used to generate captions, which are then plugged into a $[x1,y1,x2]$ input mapped into a $y*$ output for a second phase of supervised finetuning. Results in lines 7-8 (SFT+SFT2) show that this setting fails to achieve better performance. In addition, results of only train the model as correcter (namely using the initial captions in RefinedCaps pipeline instead of the first-turn generated captions to calculate loss) are reported in the last line. This setting performs worse than the proposed approach with more training data (our proposed method only needs the final caption as GT). 

\begin{table}[h]
  \centering
  \resizebox{0.99\linewidth}{!}
  {
  
  \begin{tabular}{lcccccc}
    \toprule
     Model & BLEU-4 & METEOR  & CAPTURE & O-F1 & A-F1 & Relations \\

    \midrule


     None & 29.39 & 16.59 & 57.96 & 66.47 & 52.65 & 17.57 \\
    SFT& 40.92 & 22.04 & 62.05 & 69.50 & 55.50 & 27.65 \\
    SFT+DPO$^*$& 44.49 & 23.84 & 62.51 & 70.67 & 55.60 & 27.33 \\
    SFT+Ours$^*$& \bf{44.88} & \bf{25.18} & \bf{63.34} & \bf{71.63} & \bf{57.67} & \bf{30.51} \\
    SFT+DiscriTune& 30.95 & 19.56 & 60.37 & 67.48 & 56.21 & 25.27 \\
    SFT+RL(1turn)& 41.25 & 22.91 & 63.12 & 70.57 & 57.57 & 28.42 \\
    SFT+SFT2 & 41.76 & 22.61 & 61.97 & 69.43 & 55.18 & 29.39 \\
     SFT+SFT2* & 40.19 & 22.09 & 62.03 & 70.00 & 56.02 & 29.79 \\
    SFT+Correction$^*$& 43.26 & 23.42 & 62.79 & 70.56 & 56.58 & 28.54 \\
    
    \bottomrule
  \end{tabular}
  }
  \vspace{-0.1cm}
  \caption{Experimental results with more different methods. All experiments are based on Qwen2-VL-7B and DOCCI500.}
  \label{tab:supp_ablation}
\end{table}

\begin{figure*}[h]
  \centering
  \includegraphics[width=0.99\linewidth]{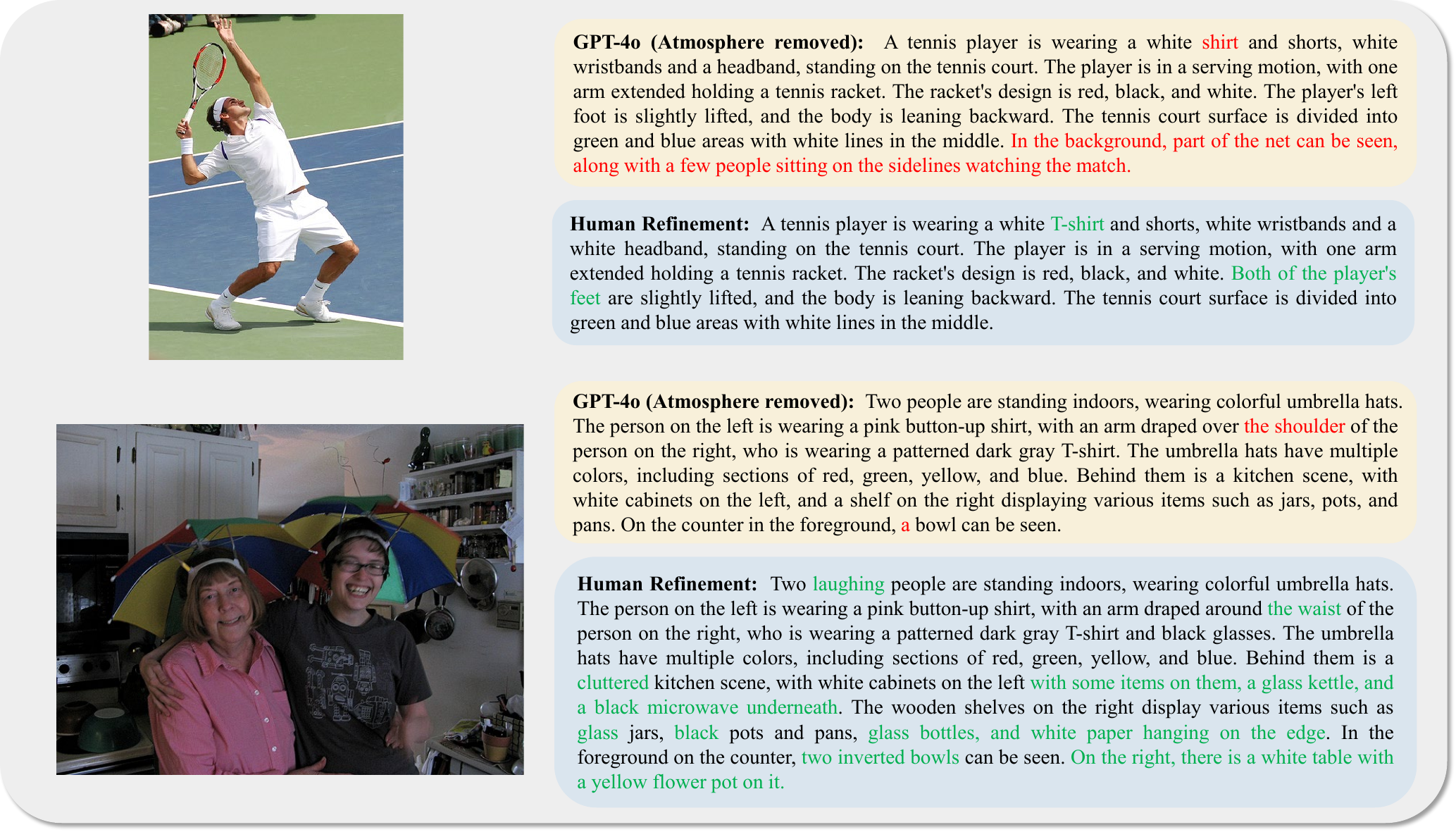}

   \caption{Visualization of human refinement samples in the RefinedCaps dataset. The red annotations represent description errors, and the green annotations represent the additional detail descriptions omitted by previous captions. Human annotators made meaningful improvements to enhance caption accuracy and completeness.}
   \label{fig:refinedcaps}
\end{figure*}

\subsection{Comparisons with more Models}

We evaluate InternVL2-8B, ShareCaptioner, Gemini-1.5 and Claude-3.7 for image captioning and compare the results on DOCCI500 in~\cref{tab:supp_models}. Closed-source models perform better than Qwen2-VL, but after training with proposed method, the baseline model can perform better.

\begin{table}[h]
\vspace{-0.3cm}
  \centering
  \renewcommand{\arraystretch}{0.85}
  \resizebox{0.99\linewidth}{!}
  {
  
  \begin{tabular}{lcccccc}
    \toprule
     Model & BLEU-4 & METEOR & CAPTURE & O-F1 & A-F1 & Relations \\

    \midrule


     Qwen2-VL-7B & 29.39 & 16.59 & 57.96 & 66.47 & 52.65 & 17.57 \\
     InternVL2-8B & 28.70 & 16.92 & 58.30 & 66.69 & 53.41 & 16.85 \\
     ShareCaptioner& 39.09 & 23.05 & 57.90 & 66.05 & 52.27 & 19.47\\
     Gemini-1.5 & 24.70 & 16.25 & 60.34 & 68.17 & 55.69 & 28.90 \\
     Claude-3.7 & 41.36 & 20.17 & 61.02 & 69.48 & 55.57 & 28.78 \\
     Qwen2-VL-7B+SFT+Ours* & \bf{44.88} & \bf{25.18} & \bf{63.34} & \bf{71.63} & \bf{57.67} & \bf{30.51} \\

    \bottomrule
  \end{tabular}
  }
  \caption{Experimental results of more models on DOCCI500.}
  \label{tab:supp_models}
\end{table}

\section{Visualization Examples}

\subsection{Annotated examples from RefinedCaps}

To better illustrate the annotation process and the quality of the RefinedCaps dataset, we present sample cases with captions before and after human refinement in~\cref{fig:refinedcaps}. These examples show that human annotators made meaningful improvements to enhance caption accuracy and completeness.

\subsection{Comparisons among Methods}
\begin{figure*}[h]
  \centering
  \includegraphics[width=\linewidth]{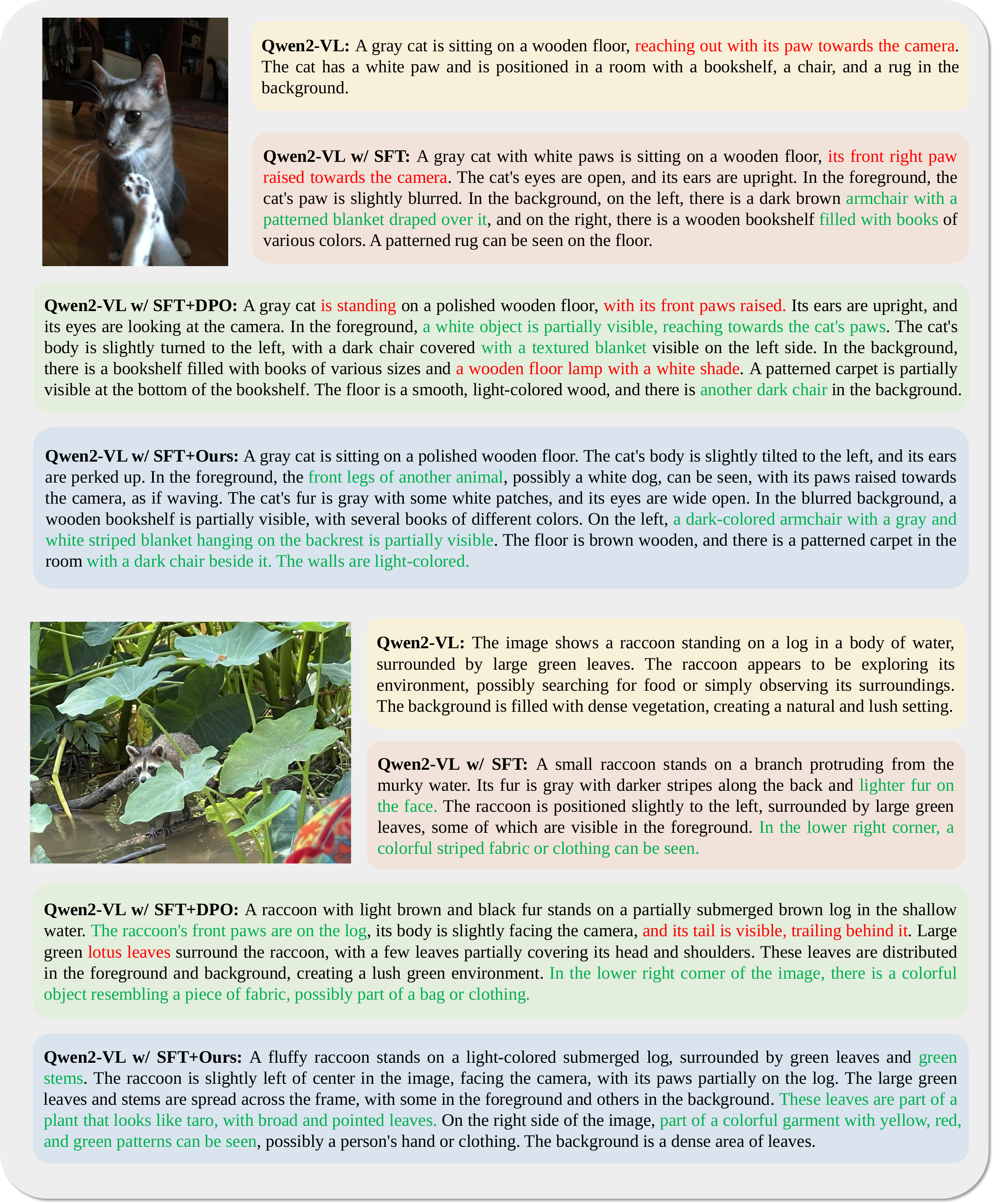}

   \caption{Additional qualitative results of baseline and three post-training approaches. The red annotations represent description errors, and the green annotations represent the additional detail descriptions omitted by previous captions. Our proposed method can reach more details while maintaining the precision of caption compared with baseline, SFT and DPO.}
   \label{fig:compare2}
\end{figure*}

To directly show the difference among different post-training methods, we visualize the caption results of baseline, SFT, SFT+DPO and SFT+ours based on Qwen2-VL-7B. The qualitative results can be referred to in~\cref{fig:compare2}. It can be seen that captions of baseline models may miss some details, and our proposed SC-Captioner can describe more details while maintaining caption precision compared with SFT and DPO.

\subsection{Self-Correction Effect}
\begin{figure*}[h]
  \centering
  \includegraphics[width=\linewidth]{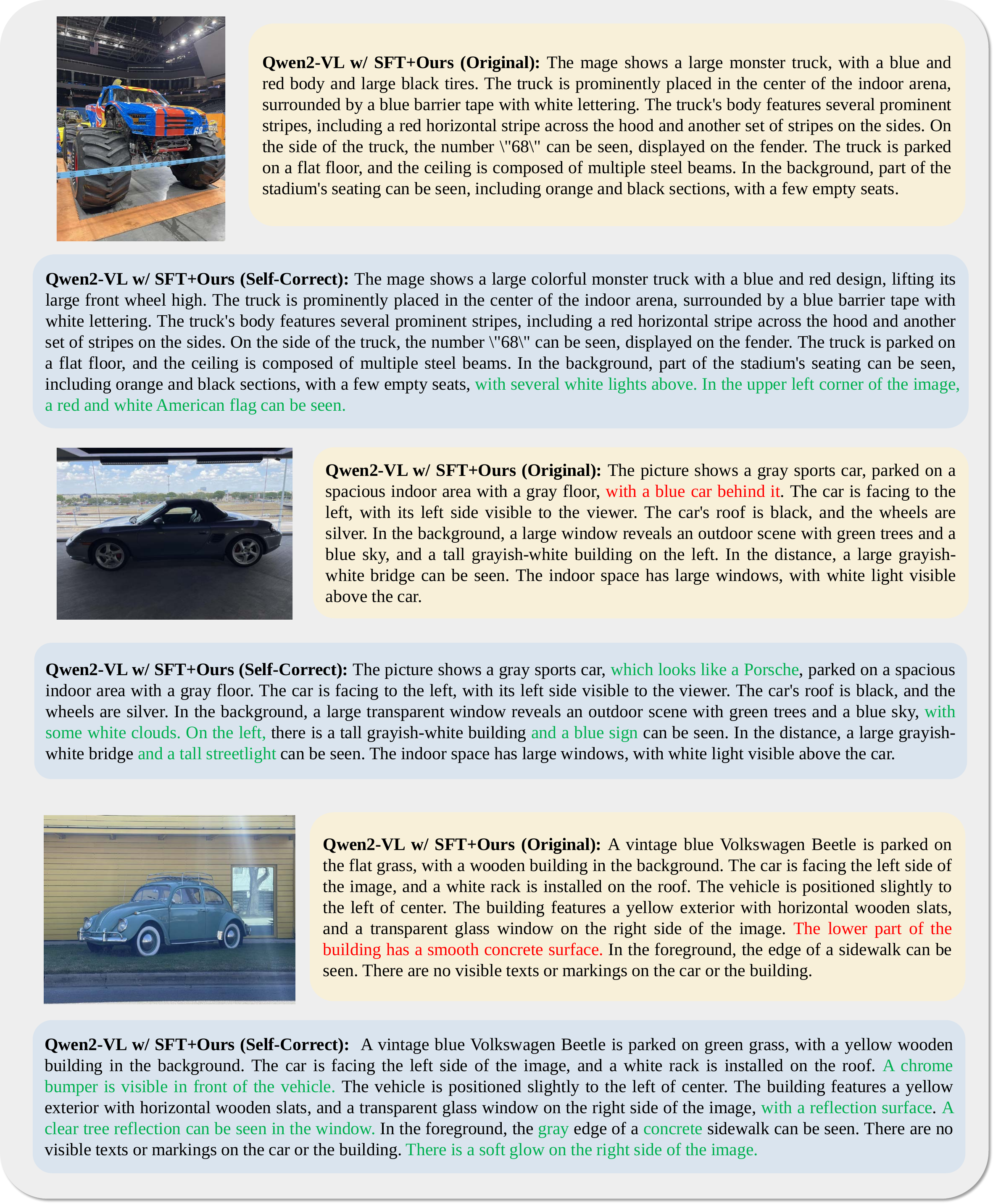}

   \caption{Qualitative results of initial and self-corrected captions. The red annotations represent deleted descriptions, and the green annotations represent the added descriptions during self-correction. The self-correcting process can delete incorrect descriptions and add more details.}
   \label{fig:compare3}
\end{figure*}

To intuitively display the self-correction effect of our proposed SC-Captioner, we visualize the initial and self-corrected captions in~\cref{fig:compare3}. It is shown that some incorrect descriptions are removed and some missing details are added during the self-correcting process.

\section{Limitations and Future Work}
The proposed framework requires ground-truth captions in the whole training process, adding to the data collection and annotation cost. Because the training process including generating annotations for two rounds, the training time is relatively long. Reinforcement learning methods that can self-explore without supervisions of ground-truth captions or even generating process may be directions for future research.  Also, more metrics including reference-free ones like~\cite{hessel2021clipscore, sarto2023pacs} and LLM-assisted ones like~\cite{lee2024fleur, ye2025dcscore} can be tested and compared in the future.

\end{document}